\documentclass[10pt]{article}

\usepackage[preprint]{icmlarxiv}

\usepackage[utf8]{inputenc}
\usepackage[T1]{fontenc}
\usepackage{amsmath,amsfonts,amssymb}
\usepackage{graphicx}
\usepackage{booktabs}
\usepackage{multirow}
\usepackage{xcolor}
\usepackage{enumitem}
\usepackage{hyperref}
\usepackage{colortbl}
\usepackage{makecell}
\usepackage{array}

\setcitestyle{numbers,square,comma,sort&compress}

\setlist[itemize]{leftmargin=*, topsep=2pt, itemsep=2pt, parsep=1pt}
\setlist[enumerate]{leftmargin=*, topsep=2pt, itemsep=2pt, parsep=1pt}

\definecolor{cellgreen}{RGB}{198,239,206}
\definecolor{cellamber}{RGB}{255,235,156}
\definecolor{cellred}{RGB}{255,199,206}

\newcommand{\cg}[1]{\cellcolor{cellgreen}#1}
\newcommand{\ca}[1]{\cellcolor{cellamber}#1}
\newcommand{\cd}[1]{\cellcolor{cellred}#1}

\begin{document}

\twocolumn[{%
\icmltitle{Quantifying Explanation Consistency: The C-Score Metric for
CAM-Based Explainability in Medical Image Classification}

\begin{icmlauthorlist}
  \icmlauthor{Kabilan Elangovan}{sgh,seri}
  \icmlauthor{Daniel Ting}{sgh,seri}
\end{icmlauthorlist}

\begin{center}
\small
\textsuperscript{1}Singapore Health Services, Singapore\
\textsuperscript{2}Singapore Eye Research Institute, Singapore\
\end{center}

\begin{abstract}
Class Activation Mapping (CAM) methods are widely used to generate visual
explanations for deep learning classifiers in medical imaging.  However,
existing evaluation frameworks assess whether explanations are
\emph{correct} measured by localisation fidelity against radiologist
annotations rather than whether they are \emph{consistent}: whether
the model applies the same spatial reasoning strategy across different
patients with the same pathology.  We propose the \textbf{C-Score}
(Consistency Score), a confidence-weighted, annotation-free metric that
quantifies intra-class explanation reproducibility via intensity-emphasised
pairwise soft-IoU across correctly classified instances.  We evaluate six
CAM techniques GradCAM, GradCAM++, LayerCAM, EigenCAM, ScoreCAM,
and MS-GradCAM++ across three CNN architectures (DenseNet201,
InceptionV3, ResNet50V2) over thirty training epochs on the Kermany chest
X-ray dataset, covering transfer learning and fine-tuning phases.  We
identify three distinct mechanisms of AUC--consistency dissociation
invisible to standard classification metrics: threshold-mediated gold-list
collapse, technique-specific attribution collapse at peak AUC, and
class-level consistency masking in global aggregation.  C-Score provides an
early-warning signal of impending model instability ScoreCAM
deterioration on ResNet50V2 is detectable one full checkpoint before
catastrophic AUC collapse and yields architecture-specific clinical
deployment recommendations grounded in explanation quality rather than
predictive ranking alone.
\end{abstract}
\vspace{0.2em}%
}]


\section{Introduction}
\label{sec:intro}

\subsection{The Clinical Deployment Gap in Medical AI}

Deep learning models for medical image analysis have achieved, and in some
domains exceeded, expert-level discriminative performance \cite{gulshan2016diabetic,esteva2017dermatologist}.
Convolutional neural networks trained on chest X-ray datasets now achieve
AUC values exceeding 0.99 for pneumonia detection \cite{kermany2018identifying},
and similar performance has been reported across ophthalmology,
dermatology, and radiology \cite{rajpurkar2017chexnet}.

However, a fundamental tension exists between classification performance
and clinical trustworthiness.  High AUC certifies that a model correctly
ranks pathological cases above normal ones in terms of predicted
probability.  It does not certify what image features the model is using
to make that ranking, nor whether those features correspond to genuine
pathological findings.  The ``Clever Hans'' phenomenon --- where models
exploit dataset-specific shortcuts rather than genuine pathological
features --- is now well-documented across medical imaging domains
\cite{degrave2021ai,zech2018pneumonia,oakden2020hidden}.  Chest X-ray classifiers have been shown to leverage equipment markers, 
patient positioning artefacts, and pacemaker presence as proxy discriminators 
\cite{winkler2019association,roberts2021common,obermeyer2019dissecting}. 
Moreover, recent perspectives highlight that AI often relies on ``subvisual'' 
or ``nonvisual'' statistical patterns---features mathematically present 
but imperceptible to radiologists---creating a fundamental verification gap 
where physicians cannot visually confirm the model's reasoning \cite{mcleod2026distinct}.

\subsection{The Reliability Problem: CAM Methods Fail Basic Sanity Checks}

Class Activation Mapping methods were introduced to address the deployment
trust gap by making CNN spatial reasoning visible and auditable
\cite{selvaraju2017gradcam}.  However, a series of foundational studies has
revealed that CAM methods themselves suffer from reliability failures that
are independent of --- and invisible to --- classification performance
metrics.

\textbf{Model Parameter Randomization} \cite{adebayo2018sanity}: Several
widely used explanation methods produce nearly identical heatmaps for
a fully trained model and for a model with randomly re-initialised weights.
If an explanation method cannot distinguish between a trained model and a
random one, it is not measuring the model's learned reasoning.

\textbf{Input Invariance Problem} \cite{kindermans2019reliability}: A
constant shift applied to all input pixels --- a transformation with
exactly zero effect on model predictions --- produces completely different
saliency attributions in many popular methods.

\textbf{Interpretation Fragility} \cite{ghorbani2019interpretation}:
Imperceptible adversarial perturbations bounded by
$L_\infty{=}2$ --- leaving model predictions unchanged --- can redirect
spatial attention to arbitrary image regions.

\textbf{Faithfulness vs.\ Visual Appeal} \cite{draelos2020hirescam}:
GradCAM's spatial pooling of gradients causes it to highlight regions
larger than those the model actually uses, creating visually appealing but
spatially imprecise attributions.

\subsection{Gaps in Current CAM Evaluation Approaches}

The recognition of CAM reliability failures has stimulated research into
quantitative evaluation frameworks, yet four systematic gaps remain.

\emph{First}, perturbation-based metrics (deletion and insertion curves)
create out-of-distribution masked inputs and conflate attribution quality
with model behaviour on corrupted data \cite{samek2017evaluating,tomsett2020sanity}.

\emph{Second}, localisation metrics compare CAM-predicted attention regions
to ground-truth annotations.  The pointing game and IoU-with-bounding-box
require expensive per-image annotation unavailable at scale and are not
applicable across training checkpoints.

\emph{Third}, and most critically: no existing evaluation framework
addresses \emph{intra-class consistency}.  Current metrics ask whether
the model attends to the correct spatial region on a per-image basis.
The fundamentally different clinical question --- does the model consistently
apply the same visual reasoning strategy across different patients with
the same diagnosis? --- remains unanswered.

\emph{Fourth}, the disagreement problem extends to the choice of evaluation
metric itself: different faithfulness metrics select different methods as
``best'' on the same model and data \cite{krishna2022disagreement,barr2023disagreement}.

\subsection{Why Consistency is the Fundamental Clinical Requirement}

The traditional assumption has been that explainability is equivalent to
highlighting the clinically correct region of interest.  A critical
reconceptualisation follows from the reliability failures documented above.
Clinical ROI alignment is desirable, but is not the fundamental requirement
for trust.  What is fundamental is whether the model consistently applies
its learned decision strategy across similar cases.

Consistent ``wrong'' focus is addressable.  Inconsistent focus is
unpredictable.  In a clinical deployment context, a physician can learn to
interpret and mentally correct for a systematic bias in AI attention; they
cannot develop a reliable mental model of a system whose attention is
arbitrarily variable.  As Ghorbani et al.\ \cite{ghorbani2019interpretation}
noted, even if a pathology predictor is robust, a fragile interpretation 
would still be highly disconcerting if a clinician is using that 
interpretation to guide clinical decision-making. Furthermore, as 
McLeod et al.\ \cite{mcleod2026distinct} note, because deep learning 
models lack causal common sense and may diverge from human visual 
search patterns, ensuring that these divergent strategies are at least consistent across patient populations is a necessary safeguard against spurious correlations. Yet, the field currently lacks a standardized, quantifiable metric to evaluate this fundamental requirement.

\subsection{The C-Score: Contribution and Positioning}

We propose the \textbf{C-Score} (Consistency Score) as a metric that
directly addresses the intra-class consistency gap. The core thesis is
that explainability itself cannot be proven because there is no objective
ground truth for what a ``correct'' explanation looks like, but consistency
in explainability can be quantified. A model that consistently attends to
the same anatomical regions across patients with the same diagnosis
demonstrates at minimum that its explanations are reproducible. This
reproducibility is a necessary precondition for clinical interpretability.

The C-Score is formulated as a confidence-weighted, intensity-emphasised
mean pairwise soft IoU across correctly classified instances of the same
class at a given model checkpoint. The metric is designed specifically for
image classification tasks trained with image-level labels and no
pixel-level annotations, where spatial explanations are derived post hoc
using CAM-based attribution methods. Under this setting, C-Score provides
an annotation-free measure of spatial explanation consistency that can be
computed at every training epoch and applied across different CAM
techniques and CNN architectures.

This paper reports five primary contributions:
\begin{enumerate}[label=(\roman*)]
  \item Formal specification of the C-Score metric with complete
        mathematical definition, intensity emphasis rationale ($\alpha{=}2.0$),
        confidence weighting, and gold-list formation under threshold $\tau{=}0.5$.
  \item Comprehensive evaluation across six CAM techniques, three CNN
        architectures, and thirty training epochs on the Kermany chest X-ray
        dataset, revealing both intra-phase and inter-phase consistency dynamics.
  \item Identification and characterisation of three distinct mechanisms of
        AUC--consistency dissociation invisible to standard classification metrics.
  \item Empirical demonstration of C-Score as a pre-collapse monitoring signal,
        with ScoreCAM deterioration on ResNet50V2 detected one training checkpoint
        before catastrophic AUC collapse.
  \item Architecture- and technique-specific clinical deployment recommendations
        grounded in both AUC and C-Score trajectory evidence.
\end{enumerate}

\section{Related Work}
\label{sec:related}

\subsection{CAM Methods for Medical Image Explainability}

Selvaraju et al.\ \cite{selvaraju2017gradcam} introduced GradCAM as a
class-discriminative visualisation method anchored to the final
convolutional layer's gradient-weighted activation.  The method produced
substantially more useful explanations than prior pixel-wise gradient
methods and achieved wide adoption in medical imaging contexts.  Subsequent
gradient-based methods refined spatial attribution precision: GradCAM++
\cite{chattopadhay2018gradcamplusplus} introduced second-order gradient
weighting; LayerCAM \cite{jiang2021layercam} preserved full spatial
resolution through pixel-wise gradient--activation products.  Gradient-free
alternatives such as ScoreCAM \cite{wang2020scorecam} and EigenCAM
\cite{muhammad2020eigencam} avoid gradient pathologies at increased
computational cost.  Multi-scale aggregation strategies have been proposed
to balance semantic strength at deep layers with spatial resolution at
shallow layers \cite{chattopadhay2018gradcamplusplus}.

Reviews of CAM methods in medical imaging (2024--2025) have consistently
identified the lack of standardised evaluation as the primary obstacle to
clinical adoption \cite{bhati2024survey,tang2024reviewing}.  Both
van der Velden et al.\ \cite{vandervelden2022xai} and Suara et al.\
\cite{suara2023gradcam} conclude that consistency, alongside faithfulness,
must be a primary evaluation criterion.

\subsection{Evaluation of Explanation Quality}

Perturbation-based metrics \cite{samek2017evaluating,tomsett2020sanity},
localisation metrics \cite{zhou2016cam}, and human-grounded assessments
\cite{lee2023haas} constitute the dominant evaluation paradigms.  The
ERASER benchmark \cite{hooker2019benchmark} and Quantus toolkit
\cite{hedstrom2023quantus,hedstrom2023metaquantus} provide multi-metric
evaluation but do not address intra-class consistency.  The disagreement
problem --- different faithfulness metrics yield conflicting rankings of
explanation methods --- has been documented by Krishna et al.\
\cite{krishna2022disagreement} and confirmed across multiple benchmarks
\cite{barr2023disagreement}.

\subsection{Consistency and Reproducibility in Prior Literature}

The closest existing work to C-Score is the Difference of Means (DoM)
metric proposed by Ozer et al.\ \cite{ozer2025consistent}, which measures
consistency of saliency detectors across different network architectures
by comparing mean activation maps.  DoM addresses inter-architecture
consistency.  C-Score addresses intra-class consistency: whether the
\emph{same} model consistently attends to the \emph{same} regions for
\emph{different patients} with the \emph{same diagnosis}.  C-Score
complements the Lago et al.\ \cite{lago2025xai} FDA-aligned consistency
dimension and differs from HAAS \cite{lee2023haas} in being annotation-free
and continuously computable across the training trajectory.

\section{Methodology}
\label{sec:method}

\subsection{Experimental Setup}

Experiments were conducted on the Kermany chest X-ray dataset
\cite{kermany2018identifying}, comprising 5,856 images with a test split
of 317 Normal and 855 Pneumonia images.  Three CNN architectures were
evaluated: DenseNet201 \cite{huang2017densenet} (20M parameters),
InceptionV3 \cite{szegedy2016inception} (24M parameters), and ResNet50V2
\cite{he2016resnet} (25M parameters), each initialised from ImageNet
\cite{deng2009imagenet} pretrained weights.

Training followed a deliberate two-phase protocol.  \emph{Phase 1 ---
Transfer Learning (epochs 1--20):} frozen backbone weights, classification
head trained with the Adam optimiser \cite{kingma2015adam} at
$\text{lr}=1{\times}10^{-4}$ with cosine annealing.  \emph{Phase 2 ---
Fine-Tuning (epochs 21--30):} all layers unfrozen at
$\text{lr}=1{\times}10^{-5}$ with label smoothing ($\varepsilon{=}0.1$) and
gradient clipping (norm $=1.0$).

\subsection{The C-Score: A Consistency Metric for CAM-Based Explanations}

\subsubsection{Motivation}

The dominant evaluation paradigm in medical AI --- AUC-ROC --- quantifies
discriminative ranking ability but provides no information about how the
model reaches its decisions.  A model achieving $\text{AUC}=0.99$ may do
so through spurious correlations with acquisition artefacts or demographic
proxies.  CAM methods were introduced to expose spatial reasoning patterns
\cite{selvaraju2017gradcam}.  However, existing CAM evaluations focus on
localisation fidelity and do not assess whether explanations are
reproducibly consistent across different patients with the same pathology.
C-Score fills this gap: rather than asking \emph{where} the model attends,
it asks \emph{how consistently} it attends to that location across the
clinical population.

\subsubsection{Formal Definition}

Let a CNN classifier $f(x;\theta)$ be parameterised by weights $\theta$
with sigmoid output $p = \sigma(f(x;\theta)) \in (0,1)$.  The
\emph{gold list} $G(c,\theta)$ for class $c$ at checkpoint $\theta$ is:
\begin{equation}
  G(c,\theta) = \bigl\{ i \in \mathcal{D}_{\text{test}} :
    y_i = c \;\wedge\; p_i^{(c,\theta)} \geq \tau \bigr\},
    \quad \tau = 0.5.
\end{equation}

\emph{Soft-IoU} between two normalised heatmaps $H_i, H_j \in [0,1]^{W
\times H}$:
\begin{equation}
  s_{\text{IoU}}(H_i, H_j) =
  \frac{\sum_{u,v} \min\!\bigl(H_i(u,v),\,H_j(u,v)\bigr)}
       {\sum_{u,v} \max\!\bigl(H_i(u,v),\,H_j(u,v)\bigr)}.
\end{equation}

\emph{Intensity emphasis} ($\alpha=2.0$ suppresses diffuse background):
\begin{equation}
  \hat{H}_i = H_i^{\,\alpha}, \qquad \alpha = 2.0.
\end{equation}

\emph{Confidence weighting}:
\begin{equation}
  w_i = \frac{p_i^{(c,\theta)}}{\sum_{j \in G(c,\theta)} p_j^{(c,\theta)}}.
\end{equation}

\emph{Per-class C-Score} (confidence-weighted mean pairwise soft-IoU):
\begin{equation}
  C(c,\theta,m) =
  \frac{\displaystyle\sum_{\substack{i,j \in G(c,\theta) \\ i < j}}
        (w_i + w_j)\cdot s_{\text{IoU}}(\hat{H}_i,\hat{H}_j)}
       {Z},
\end{equation}
where $Z$ is the normalisation constant over all pairs.

\emph{Global support-weighted C-Score}:
\begin{equation}
  C_{\text{global}}(\theta,m) =
  \sum_c \frac{|G(c,\theta)|}{\sum_{c'} |G(c',\theta)|}
  \cdot C(c,\theta,m).
\end{equation}

Table~\ref{tab:notation} summarises the notation.

\begin{table}[t]
\centering
\caption{C-Score notation and definitions.}
\label{tab:notation}
\begin{small}
\setlength{\tabcolsep}{4pt}
\begin{tabular}{lp{5.2cm}}
\toprule
\textbf{Symbol} & \textbf{Definition} \\
\midrule
$C(c,\theta,m)$ & C-Score for class $c$, checkpoint $\theta$, method $m$; $C \in [0,1]$ \\
$G(c,\theta)$   & Gold list: correctly classified test images for class $c$ at $\theta$ under $\tau{=}0.5$ \\
$H_i^{(c,\theta,m)}$ & Normalised CAM heatmap; $H_i \in [0,1]^{W\times H}$ \\
$p_i^{(c,\theta)}$   & Sigmoid confidence for class $c$ on image $i$ \\
$w_i$           & Confidence weight: $w_i = p_i / \sum_j p_j$ (normalised over gold list) \\
$s_{\text{IoU}}(H_i,H_j)$ & Soft-IoU: $\sum \min(H_i,H_j) / \sum \max(H_i,H_j)$, element-wise \\
$\hat{H}_i = H_i^{\,\alpha}$ & Intensity emphasis; $\alpha=2.0$ \\
$C_{\text{global}}(\theta,m)$ & Support-weighted global C-Score across classes \\
$\tau = 0.5$    & Classification threshold for gold list membership \\
$\ell^*$        & Target layer: DenseNet201$\to$\texttt{conv5\_block32\_concat}; InceptionV3$\to$\texttt{mixed10}; ResNet50V2$\to$\texttt{conv5\_block3\_out} \\
\bottomrule
\end{tabular}
\end{small}
\end{table}

\subsubsection{Theoretical Motivation: The Gradient Flow--Consistency Connection}

We hypothesize that during transfer learning, frozen backbone weights restrict gradient signals to flow only through the classification head, resulting in early explanations that lack consistency. At the target layer $\ell^{*}$, gradients therefore reflect a domain-incomplete pathway. Consequently, two pneumonia patients activating different ImageNet-derived features are likely to produce uncorrelated gradients, predicting a low initial C-Score for gradient-based attribution methods.

In contrast, fine-tuning is expected to propagate chest X-ray–specific gradients through the entire backbone, aligning internal activations toward domain-relevant feature axes and thereby increasing cross-patient heatmap similarity. EigenCAM, which is entirely gradient-free, should theoretically remain unaffected by this phenomenon.

\subsubsection{Relationship to Existing Frameworks}

C-Score complements the Lago et al.\ \cite{lago2025xai} FDA-aligned
consistency dimension and differs from HAAS \cite{lee2023haas} in being
annotation-free and continuously computable.  It operationalises a
property not captured by deletion/insertion metrics, pointing game, or
IoU with bounding box, each of which requires per-image evaluation rather
than intra-class population assessment.

\subsection{CAM Methods Evaluated}

Six CAM techniques were evaluated, applied to fixed target layers
(DenseNet201: \texttt{conv5\_block32\_concat}; InceptionV3: \texttt{mixed10};
ResNet50V2: \texttt{conv5\_block3\_out}).

\textbf{GradCAM} \cite{selvaraju2017gradcam}: Weighted linear combination
of feature maps using globally average-pooled class-specific gradients:
$\alpha_k^c = \frac{1}{Z}\sum_{i,j} \partial y^c / \partial A^k_{ij}$;
$L^c = \text{ReLU}(\sum_k \alpha_k^c \cdot A^k)$.  Reference baseline method.
Global pooling discards spatial gradient structure, producing coarse
attribution blobs.

\textbf{GradCAM++} \cite{chattopadhay2018gradcamplusplus}: Spatially
resolved second-order gradient weighting:
$\alpha_k^c = \sum_{i,j} w_k^{ij,c} \cdot \text{ReLU}(\partial y^c / \partial A^k_{ij})$.
More effective for multi-instance patterns; consistently outperforms
GradCAM in C-Score due to finer spatial discrimination.

\textbf{LayerCAM} \cite{jiang2021layercam}: Element-wise product of
positive gradients and activations:
$L^c = \text{ReLU}(\partial y^c / \partial A) \odot A$.  Retains full
spatial resolution, producing compact attribution footprints.  Achieves
C-Score values comparable to GradCAM++ across all architectures.

\textbf{EigenCAM} \cite{muhammad2020eigencam}: SVD of the activation
tensor $A_{\text{flat}} = U\Sigma V^\top$; first right singular vector
$v_1$ reshaped to spatial attribution.  Entirely gradient-free:
insensitive to gradient noise, saturation, and sparsity.  Achieves
uniquely stable C-Score throughout transfer learning (DenseNet201:
0.635 at E1, 0.635 at E20; InceptionV3: 0.758 at E1, 0.756 at E20).

\textbf{ScoreCAM} \cite{wang2020scorecam}: Per-channel input masking with
forward-pass class probability measurement:
$w^k = f^c(X \odot M^k) - f^c(X_{\text{baseline}})$.  Gradient-free;
$O(C)$ forward passes; subsampling at \texttt{max\_N=32} applied.  Its
collapse trajectory on ResNet50V2 provides the study's clearest
pre-collapse early-warning signal.

\textbf{MS-GradCAM++ (Multi-Scale GradCAM++):} To address the trade-off
between semantic strength at deep layers and spatial resolution at shallow
layers, we implement a multi-scale aggregation strategy.  We compute
GradCAM++ heatmaps at $K$ distinct points in the feature hierarchy (e.g.,
for DenseNet201: \texttt{conv3\_block12}, \texttt{pool4}, and
\texttt{conv5\_block32}) and compute their pixel-wise arithmetic mean:
\begin{equation}
  L^c_{\text{MS}} = \frac{1}{K} \sum_{\ell \in \mathcal{L}}
    \uparrow\!\bigl(L^c_{\text{GradCAM++}}(\ell)\bigr),
\end{equation}
where $\uparrow(\cdot)$ denotes bilinear upsampling to the input
resolution.  This approach stabilises the attribution map by smoothing
layer-specific gradient noise, providing robustness against severe
architectural instabilities (ResNet50V2 net $\Delta = {-}0.098$ vs.\
ScoreCAM $\Delta = {-}0.612$), at the cost of diluted semantic specificity
in fully fine-tuned models.

\subsection{Evaluation Protocol}

C-Score was computed at seven checkpoints per architecture (epochs 1, 5,
10, 15, 20, 25, 30) from per-architecture trajectory CSV logs.  Test AUC
and accuracy were extracted from \texttt{epoch\_metrics.csv} training logs
for all three architectures across all 30 epochs.  The gold list was anchored
to the epoch-30 reference model at $\tau=0.5$, yielding 317 Normal and
855 Pneumonia test-set images for binary evaluation.

\section{Results}
\label{sec:results}

Complete result tables and visualisations are provided in
Appendix~\ref{app:results}.

\subsection{Full Classification Performance Trajectory}

Table~\ref{tab:auc} presents the complete per-epoch AUC and test accuracy
for all three architectures across all 30 training epochs.
DenseNet201 achieves monotonically increasing AUC in transfer learning
(0.9184$\to$0.9902, E1$\to$E20), suffers boundary-reorganisation accuracy
collapse at E21--23 (27--28\% accuracy, AUC $>$0.984), recovers by E24,
and peaks at E30 (AUC $=$ 0.9945).  InceptionV3 grows more slowly through
transfer learning and undergoes a sharper accuracy collapse at E23 (28.75\%)
before resolving at E25 (96.76\%).  ResNet50V2 experiences catastrophic mode
collapse at E23 (AUC $=$ 0.0287) and again at E30 (AUC $=$ 0.1034),
representing a complete failure of fine-tuning stability.

\subsection{Global Weighted C-Score Trajectory}

\begin{figure}[!htbp]
\centering
\includegraphics[width=\textwidth]{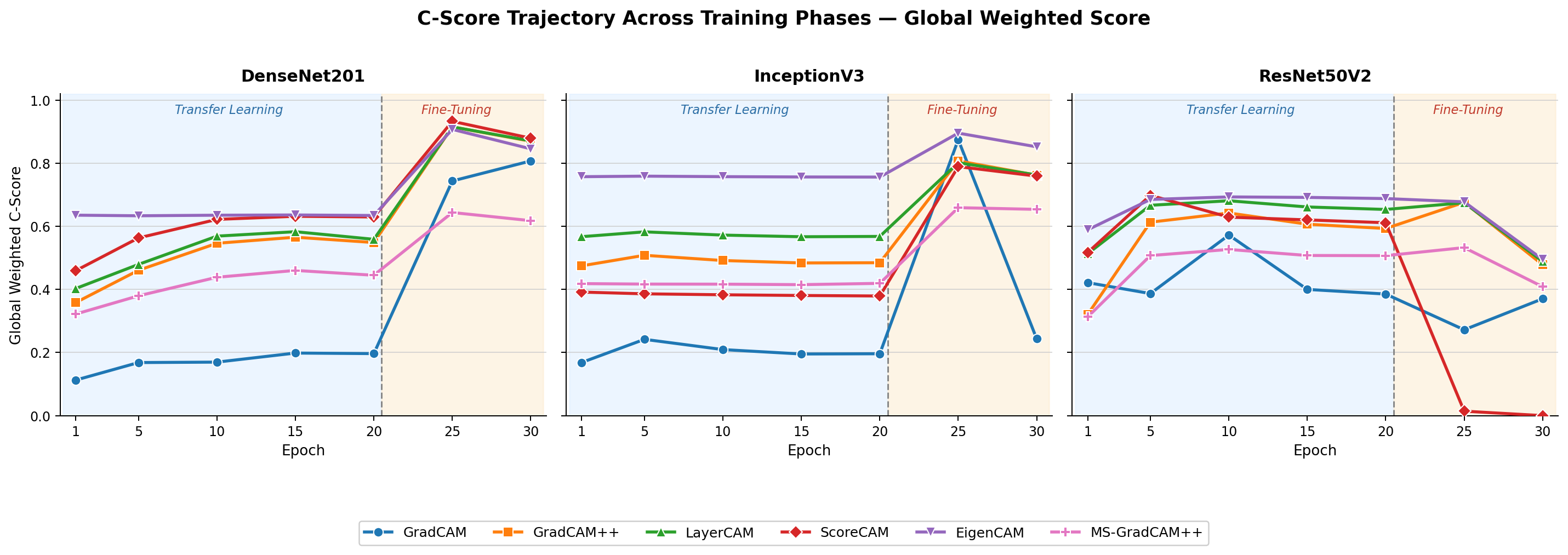}
\caption{Global weighted C-Score trajectory across training phases.}
\label{fig:global_cscore}
\end{figure}

\begin{figure}[!htbp]
\centering
\includegraphics[width=\textwidth]{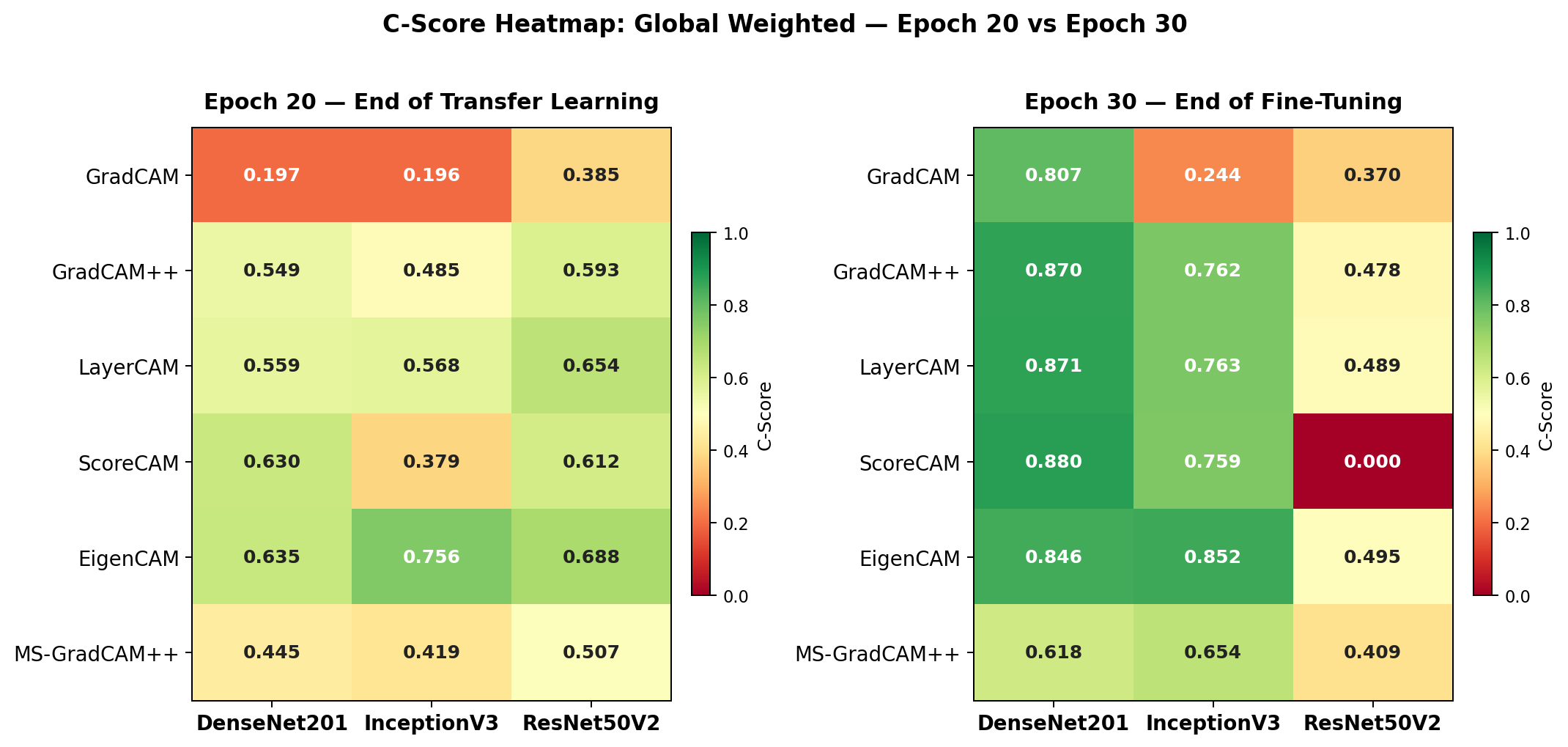}
\caption{C-Score heatmap comparison at transfer learning end (E20) and fine-tuning end (E30).}
\label{fig:heatmap}
\end{figure}

Tables~\ref{tab:cscore_densenet}--\ref{tab:cscore_resnet} present the
global weighted C-Score at seven checkpoints for all three architectures.
Figure~\ref{fig:global_cscore} visualises the trajectories;
Figure~\ref{fig:heatmap} compares heatmaps at E20 vs.\ E30.

\subsubsection{DenseNet201 --- Systematic Fine-Tuning Improvement}

Every technique shows positive net change E20$\to$E30.  GradCAM achieves
the largest absolute gain (+0.610: 0.197$\to$0.807), consistent with the
gradient flow--consistency coupling hypothesis.  GradCAM++, LayerCAM, and
ScoreCAM converge to 0.870--0.880 at E30, indicating method-equivalence
for a well-trained stable model.  EigenCAM maintains high stability
throughout ($\approx$0.635) and further improves at fine-tuning (0.846).

\subsubsection{InceptionV3 --- Transfer-Learning Stasis, Fine-Tuning Resolution}

All gradient-based techniques plateau during transfer learning (GradCAM:
0.169$\to$0.196; ScoreCAM: 0.392$\to$0.379), reflecting gradient
dispersion across parallel inception modules.  EigenCAM maintains stability
throughout (0.758$\to$0.756).  Fine-tuning produces a dramatic step-change
but with non-monotonic GradCAM behaviour (0.875 at E25, collapsing to 0.244
at E30).

\subsubsection{ResNet50V2 --- Fine-Tuning Degradation Preceding Mode Collapse}

\begin{figure}[!htbp]
\centering
\includegraphics[width=\textwidth]{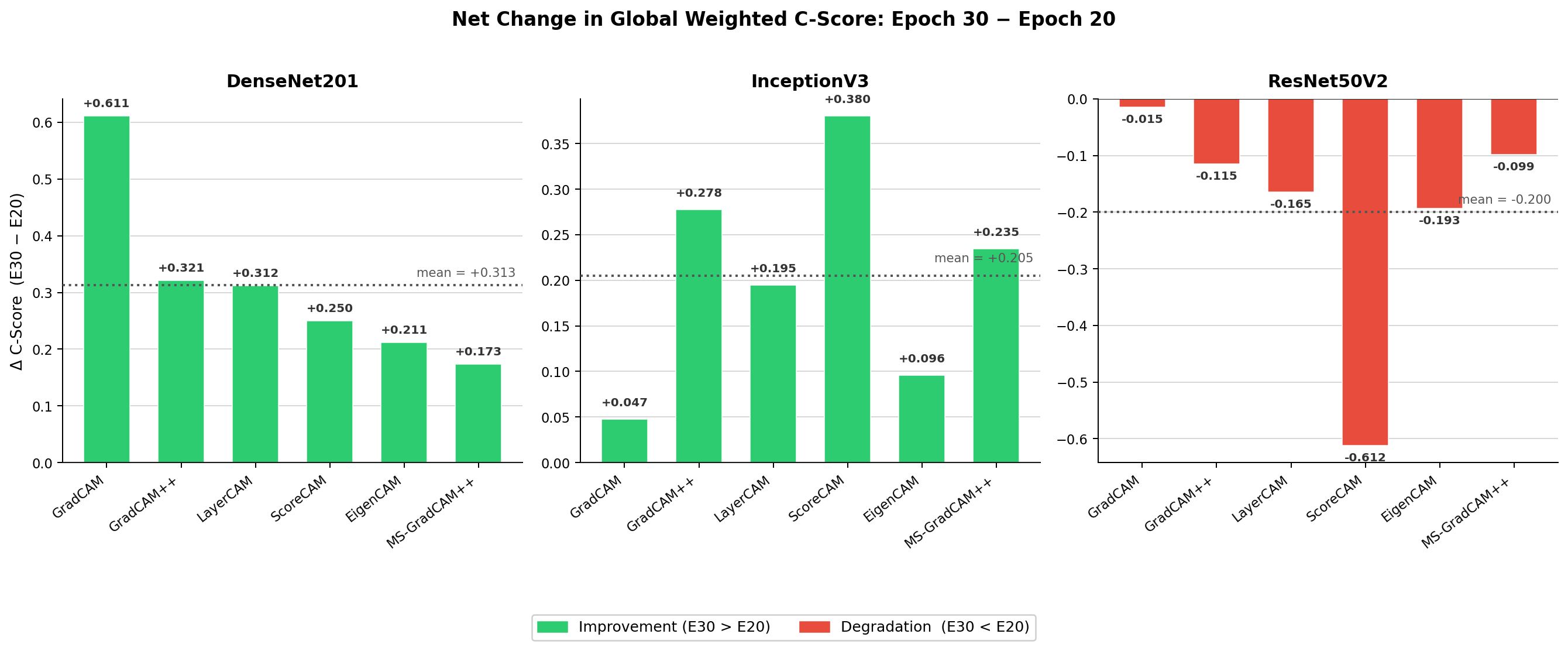}
\caption{Net C-Score change (E30$-$E20) by architecture and method.}
\label{fig:net_change}
\end{figure}

All techniques show net negative E20$\to$E30 change (Figure~\ref{fig:net_change}).
ScoreCAM provides the clearest diagnostic signal: $0.612$ (E20) $\to$
$0.014$ (E25) $\to$ $0.000$ (E30), detectable one full checkpoint before
the AUC collapse at E30.  This demonstrates C-Score's potential as an
annotation-free deployment monitoring metric for production systems.

\subsection{Per-Class C-Score: All Architectures}

\begin{figure}[!htbp]
\centering
\includegraphics[width=\textwidth]{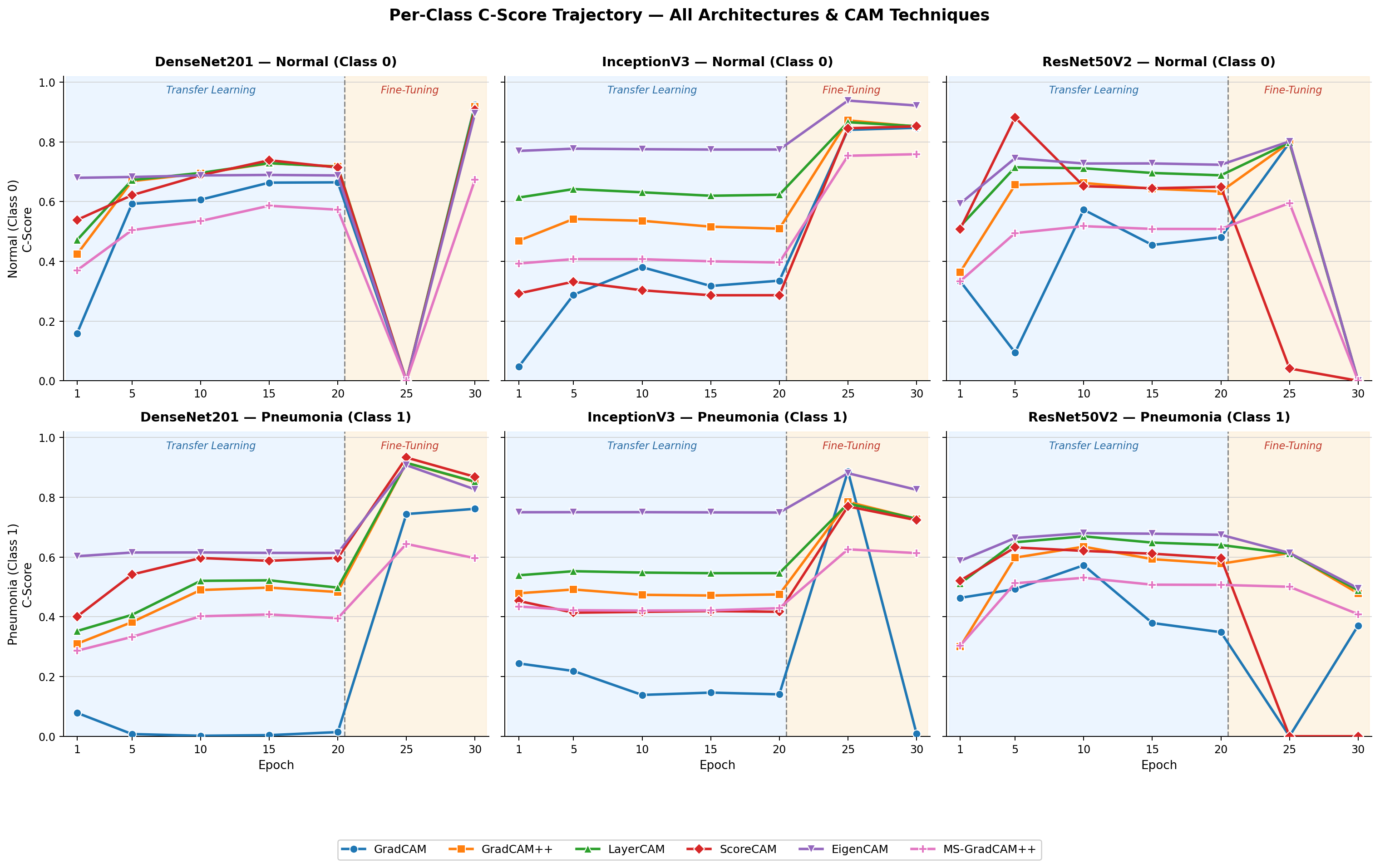}
\caption{Per-class C-Score trajectory by architecture.}
\label{fig:perclass}
\end{figure}

Tables~\ref{tab:perclass_densenet}--\ref{tab:perclass_resnet} present
per-class C-Score at seven checkpoints; Figure~\ref{fig:perclass}
visualises the trajectories.

\subsubsection{The GradCAM Class Gap: DenseNet201 Transfer Learning}

GradCAM on DenseNet201 produces a Pneumonia C-Score that remains near-zero
throughout the entire transfer learning phase (0.078 at E1, 0.007 at E5,
0.002 at E10, 0.004 at E15, 0.014 at E20) while Normal C-Score reaches
0.664 --- a class gap of 0.650 at equal AUC $=$ 0.9902.  The model
classifies pneumonia cases correctly but attends to entirely different
regions for each patient.  No classification metric --- AUC, accuracy, or
F1 --- provides any signal of this class-level explanation failure.

\subsection{The Fine-Tuning Paradox: Three Mechanisms of AUC--Consistency Dissociation}

\textbf{Mechanism 1 --- Threshold-mediated gold-list collapse.}
DenseNet201 epochs 21--23: accuracy $\approx$ 27\%, AUC $>$ 0.984.
C-Score correctly reports zero explanation consistency for the Normal class
because no Normal image passes $\tau=0.5$.  AUC conceals this operational
failure entirely.  Deploying at E22 (AUC $=$ 0.9852) would mean deploying
a model that generates empty explanations for the Normal class.

\textbf{Mechanism 2 --- Technique-specific attribution collapse at peak AUC.}
ScoreCAM on ResNet50V2 at E30: AUC $=$ 0.1034 (collapsed), C-Score $=$
0.000.  More critically, ScoreCAM C-Score degrades to 0.014 at E25 while
AUC remains at 0.9902 --- the consistency failure precedes the
classification failure by a full checkpoint.  No classification metric
provides any signal of this impending collapse.

\textbf{Mechanism 3 --- Class-level consistency masking in global metrics.}
DenseNet201 GradCAM at E20: global C-Score $=$ 0.197, Normal $=$ 0.664,
Pneumonia $=$ 0.014.  Global aggregation masks near-total explanation
failure for the clinically critical Pneumonia class.  Per-class reporting
is essential.

These three mechanisms collectively demonstrate that neither AUC nor global
C-Score alone suffices for clinical AI quality assurance.  The minimum
acceptable evaluation framework is per-class C-Score tracked across the
full training trajectory, using multiple CAM techniques, with explicit
reporting of gold-list population at each checkpoint.

\section{Discussion}
\label{sec:discussion}

\subsection{Architecture Recommendations}

DenseNet201 is recommended for clinical deployment: highest stable AUC
(0.9945), broadest consistency improvement across fine-tuning (average
$\Delta = +0.340$ across six techniques), smallest class gap, and no
catastrophic failures.  InceptionV3 achieves marginally higher AUC
(0.9949) at the cost of pronounced GradCAM instability between E25 and E30,
making technique selection critical if deployed.  ResNet50V2 is not
recommended for clinical use: fine-tuning consistently degrades explanation
consistency, ScoreCAM collapse preceded AUC collapse, and both classes
register zero C-Score at E30.

\subsection{C-Score as a Deployment Monitoring Tool}

The pre-collapse ScoreCAM signal on ResNet50V2 (C-Score $0.612 \to 0.014$
at E25, one checkpoint before AUC collapse at E30) demonstrates that
explanation consistency can deteriorate ahead of classification performance.
In production systems with periodic weight updates or continued learning,
C-Score monitoring provides an additional safety layer not available from
AUC monitoring alone.  The annotation-free nature of C-Score makes this
monitoring scalable across deployment environments without requiring
radiologist input at each update cycle.

\subsection{Broader Implications for Clinical AI Validation}

The three AUC--consistency dissociation mechanisms documented here have
direct regulatory implications.  The European AI Act \cite{eu2024aiact}
designates medical AI as high-risk and requires providers to demonstrate
that high-risk AI systems produce consistent and interpretable outputs.
FDA guidance \cite{fda2021samd} on AI/ML-based software as a medical device
emphasises the need for performance monitoring across the model lifecycle.
C-Score provides a concrete, continuously computable metric for both
requirements: it quantifies explanation consistency without requiring
ground-truth annotations and can be computed at any post-deployment
checkpoint using only the existing test set.  The analogy to inter-rater
reliability in clinical practice \cite{landis1977kappa} grounds C-Score
within established clinical validation methodology.

\section{Limitations}
\label{sec:limitations}

\textbf{CAM scope.} C-Score is formulated for 2D spatial heatmaps from
convolutional layers.  Not directly applicable to input-space methods
(SmoothGrad \cite{smilkov2017smoothgrad}, Integrated Gradients
\cite{sundararajan2017axiomatic}) or model-agnostic methods (SHAP
\cite{lundberg2017shap}, LIME \cite{ribeiro2016lime}).

\textbf{Layer sensitivity.} C-Score values are sensitive to target layer
depth.  Shallower layers produce diffuse activations that may affect
C-Score independently of classification quality.  Systematic layer-depth
sensitivity analysis is required for cross-architecture standardisation.

\textbf{Threshold dependency.} The $\tau=0.5$ threshold determines gold-list
membership.  Degenerate threshold-crossing regimes produce empty gold lists.
Threshold-adaptive variants and calibrated probability weighting should be
explored.

\textbf{Training log completeness.} InceptionV3 shows transient accuracy
collapse events at epochs 23 and 26 whose mechanistic origins --- gradient
explosion, learning-rate schedule artefacts, or batch variance --- were not
fully diagnosed.  Future work should instrument training with per-layer
gradient-norm logging.

\textbf{Single dataset.} Results are from one binary classification task on
one publicly available dataset.  Generalisation to multi-class settings,
CT/MRI modalities, and transformer-based architectures
\cite{tan2019efficientnet,liu2022convnext,dosovitskiy2021vit} requires
dedicated investigation.

\textbf{Consistency $\neq$ Correctness.} High C-Score certifies spatial
reproducibility, not clinical correctness.  A model consistently attending
to spurious correlates achieves high C-Score.  Validation against
radiologist-annotated ground-truth is necessary to confirm that consistent
explanations are also clinically faithful \cite{larrazabal2020gender}.

\section{Conclusion}
\label{sec:conclusion}

We proposed the C-Score (Consistency Score), a confidence-weighted,
annotation-free metric for quantifying intra-class explanation
reproducibility of CAM-based methods in medical image classification.
Evaluated across six CAM techniques, three CNN architectures, and thirty
training epochs on the Kermany chest X-ray dataset, C-Score revealed three
distinct mechanisms of AUC--consistency dissociation invisible to standard
classification metrics.  DenseNet201 demonstrated the most favourable
profile for clinical deployment across both dimensions.  C-Score provides
a practical, continuously computable quality assurance signal that
complements existing evaluation frameworks and satisfies the consistency
requirements emerging from regulatory guidance on clinical AI.


\bibliographystyle{unsrtnat}
\bibliography{references}

\clearpage
\onecolumn

\appendix
\section{Result Tables and Figures}
\label{app:results}

\begin{table}[!htbp]
\centering
\caption{Complete 30-epoch test AUC and accuracy for all three architectures.
TL = Transfer Learning (epochs 1--20); FT = Fine-Tuning (epochs 21--30).
\colorbox{cellred}{Red}: AUC $<0.50$ (mode collapse).
\colorbox{cellamber}{Amber}: accuracy $<50\%$ with intact AUC (threshold-crossing, rankings preserved).
DenseNet201 epochs 21--23 exhibit boundary-reorganisation accuracy collapse while AUC remains high.}
\label{tab:auc}
\begin{small}
\setlength{\tabcolsep}{5pt}
\begin{tabular}{clcccccc}
\toprule
& & \multicolumn{2}{c}{\textbf{DenseNet201}} & \multicolumn{2}{c}{\textbf{InceptionV3}} & \multicolumn{2}{c}{\textbf{ResNet50V2}} \\
\cmidrule(lr){3-4}\cmidrule(lr){5-6}\cmidrule(lr){7-8}
\textbf{Ep.} & \textbf{Phase} & \textbf{AUC} & \textbf{Acc.} & \textbf{AUC} & \textbf{Acc.} & \textbf{AUC} & \textbf{Acc.} \\
\midrule
1  & TL & 0.9184 & 61.60\% & 0.8705 & 66.13\% & 0.9582 & 81.48\% \\
2  & TL & 0.9656 & 84.64\% & 0.9310 & 64.33\% & 0.9760 & 94.20\% \\
3  & TL & 0.9657 & 45.48\% & 0.9509 & 69.97\% & 0.9807 & 82.42\% \\
4  & TL & 0.9804 & 90.96\% & 0.9542 & 86.35\% & 0.9800 & 94.71\% \\
5  & TL & 0.9848 & 94.03\% & 0.9592 & 78.33\% & 0.9876 & 95.82\% \\
6  & TL & 0.9855 & 92.58\% & 0.9626 & 80.46\% & 0.9885 & 95.56\% \\
7  & TL & 0.9859 & 94.97\% & 0.9602 & 85.49\% & 0.9859 & 95.82\% \\
8  & TL & 0.9865 & 90.44\% & 0.9641 & 84.13\% & 0.9888 & 90.79\% \\
9  & TL & 0.9882 & 94.80\% & 0.9633 & 87.63\% & 0.9894 & 94.71\% \\
10 & TL & 0.9886 & 94.88\% & 0.9664 & 86.86\% & 0.9892 & 93.52\% \\
11 & TL & 0.9889 & 94.88\% & 0.9673 & 87.20\% & 0.9908 & 95.05\% \\
12 & TL & 0.9890 & 95.14\% & 0.9672 & 87.03\% & 0.9909 & 94.28\% \\
13 & TL & 0.9893 & 95.31\% & 0.9675 & 88.99\% & 0.9895 & 95.90\% \\
14 & TL & 0.9893 & 94.80\% & 0.9661 & 88.23\% & 0.9900 & 84.39\% \\
15 & TL & 0.9903 & 90.79\% & 0.9679 & 88.91\% & 0.9902 & 94.97\% \\
16 & TL & 0.9890 & 95.73\% & 0.9661 & 89.25\% & 0.9893 & 92.06\% \\
17 & TL & 0.9901 & 95.73\% & 0.9678 & 87.88\% & 0.9901 & 92.49\% \\
18 & TL & 0.9902 & 95.90\% & 0.9676 & 87.46\% & 0.9891 & 94.62\% \\
19 & TL & 0.9899 & 95.14\% & 0.9684 & 88.82\% & 0.9888 & 95.14\% \\
20 & TL & 0.9902 & 94.80\% & 0.9680 & 89.33\% & 0.9898 & 94.20\% \\
\midrule
21 & FT & \ca{0.9842} & \ca{27.73\%} & 0.9297 & 84.30\% & 0.9876 & 79.86\% \\
22 & FT & \ca{0.9852} & \ca{27.05\%} & 0.9648 & 72.95\% & 0.9807 & 42.49\% \\
23 & FT & \ca{0.9891} & \ca{27.05\%} & \ca{0.9836} & \ca{28.75\%} & \cd{0.0287} & 70.99\% \\
24 & FT & 0.9910 & 83.70\% & 0.9892 & 69.20\% & 0.9885 & 44.45\% \\
25 & FT & \ca{0.9867} & \ca{72.95\%} & 0.9902 & 96.76\% & 0.9902 & 78.75\% \\
26 & FT & \ca{0.9844} & \ca{72.95\%} & \ca{0.9930} & \ca{38.23\%} & 0.9868 & 82.34\% \\
27 & FT & 0.9931 & 96.93\% & 0.9925 & 87.46\% & 0.9933 & 95.39\% \\
28 & FT & 0.9927 & 96.76\% & 0.9949 & 97.61\% & 0.9938 & 74.91\% \\
29 & FT & 0.9941 & 86.77\% & 0.9943 & 94.62\% & 0.9947 & 97.01\% \\
30 & FT & 0.9945 & 94.20\% & 0.9949 & 94.62\% & \cd{0.1034} & 72.95\% \\
\bottomrule
\end{tabular}
\end{small}
\end{table}

\vspace{0.5em}

\begin{table}[!htbp]
\centering
\caption{Global weighted C-Score for DenseNet201.
\colorbox{cellgreen}{Green}: C-Score $\geq 0.80$;
\colorbox{cellamber}{amber}: $< 0.30$;
\colorbox{cellred}{red}: $0.000$.
Step-change improvement at TL$\to$FT transition observable across all methods.}
\label{tab:cscore_densenet}
\begin{small}
\setlength{\tabcolsep}{5pt}
\begin{tabular}{lccccccc}
\toprule
\textbf{Technique} & \textbf{E1} & \textbf{E5} & \textbf{E10} & \textbf{E15}
  & \makecell{\textbf{E20}\\\small{(TL End)}}
  & \makecell{\textbf{E25}\\\small{(Mid-FT)}}
  & \makecell{\textbf{E30}\\\small{(FT End)}} \\
\midrule
GradCAM        & \ca{0.113} & \ca{0.168} & \ca{0.170} & \ca{0.198} & \ca{0.197} & \cg{0.744} & \cg{0.807} \\
GradCAM++      & 0.358 & 0.461 & 0.546 & 0.566 & 0.549 & \cg{0.916} & \cg{0.870} \\
LayerCAM       & 0.403 & 0.479 & 0.569 & 0.583 & 0.559 & \cg{0.915} & \cg{0.871} \\
ScoreCAM       & 0.460 & 0.563 & 0.622 & 0.632 & 0.630 & \cg{0.933} & \cg{0.880} \\
EigenCAM       & 0.635 & 0.634 & 0.635 & 0.636 & 0.635 & \cg{0.908} & \cg{0.846} \\
MS-GradCAM++   & 0.322 & 0.380 & 0.439 & 0.460 & 0.445 & 0.644 & 0.618 \\
\bottomrule
\end{tabular}
\par\vspace{3pt}
{\footnotesize $\Delta$ (E30$-$E20): +0.610 (GradCAM); +0.321 (GradCAM++); +0.312 (LayerCAM); +0.250 (ScoreCAM); +0.211 (EigenCAM); \textbf{avg: +0.267}}
\end{small}
\end{table}

\begin{table}[!htbp]
\centering
\caption{Global weighted C-Score for InceptionV3.}
\label{tab:cscore_inceptionv3}
\begin{small}
\setlength{\tabcolsep}{5pt}
\begin{tabular}{lccccccc}
\toprule
\textbf{Technique} & \textbf{E1} & \textbf{E5} & \textbf{E10} & \textbf{E15}
  & \makecell{\textbf{E20}\\\small{(TL End)}}
  & \makecell{\textbf{E25}\\\small{(Mid-FT)}}
  & \makecell{\textbf{E30}\\\small{(FT End)}} \\
\midrule
GradCAM        & \ca{0.169} & \ca{0.242} & \ca{0.209} & \ca{0.195} & \ca{0.196} & \cg{0.875} & \ca{0.244} \\
GradCAM++      & 0.475 & 0.508 & 0.492 & 0.484 & 0.485 & \cg{0.808} & 0.762 \\
LayerCAM       & 0.567 & 0.583 & 0.572 & 0.567 & 0.568 & \cg{0.802} & 0.763 \\
ScoreCAM       & 0.392 & 0.386 & 0.383 & 0.381 & 0.379 & 0.790 & 0.759 \\
EigenCAM       & 0.758 & 0.759 & 0.758 & 0.757 & 0.756 & \cg{0.896} & \cg{0.852} \\
MS-GradCAM++   & 0.419 & 0.417 & 0.417 & 0.415 & 0.419 & 0.659 & 0.654 \\
\bottomrule
\end{tabular}
\par\vspace{3pt}
{\footnotesize $\Delta$ (E30$-$E20): +0.048 (GradCAM); +0.277 (GradCAM++); +0.195 (LayerCAM); +0.380 (ScoreCAM); +0.096 (EigenCAM); \textbf{avg: +0.158}}
\end{small}
\end{table}

\begin{table}[!htbp]
\centering
\caption{Global weighted C-Score for ResNet50V2.
Net E20$\to$E30 change negative for majority of techniques.
ScoreCAM trajectory $0.612 \to 0.014 \to 0.000$ provides pre-collapse early-warning signal
one checkpoint before AUC collapse at E30.}
\label{tab:cscore_resnet}
\begin{small}
\setlength{\tabcolsep}{5pt}
\begin{tabular}{lccccccc}
\toprule
\textbf{Technique} & \textbf{E1} & \textbf{E5} & \textbf{E10} & \textbf{E15}
  & \makecell{\textbf{E20}\\\small{(TL End)}}
  & \makecell{\textbf{E25}\\\small{(Mid-FT)}}
  & \makecell{\textbf{E30}\\\small{(FT End)}} \\
\midrule
GradCAM        & 0.422 & 0.387 & 0.573 & 0.400 & 0.385 & \ca{0.272} & 0.370 \\
GradCAM++      & \ca{0.320} & 0.613 & 0.642 & 0.607 & 0.593 & 0.676 & 0.478 \\
LayerCAM       & 0.513 & 0.667 & 0.681 & 0.662 & 0.654 & 0.675 & 0.489 \\
ScoreCAM       & 0.517 & 0.698 & 0.629 & 0.621 & 0.612 & \ca{0.014} & \cd{0.000} \\
EigenCAM       & 0.589 & 0.685 & 0.693 & 0.692 & 0.688 & 0.678 & 0.495 \\
MS-GradCAM++   & \ca{0.313} & 0.508 & 0.527 & 0.508 & 0.507 & 0.533 & 0.409 \\
\bottomrule
\end{tabular}
\par\vspace{3pt}
{\footnotesize $\Delta$ (E30$-$E20): $-$0.015 (GradCAM); $-$0.115 (GradCAM++); $-$0.165 (LayerCAM); $-$0.612 (ScoreCAM); $-$0.193 (EigenCAM); \textbf{avg: $-$0.162}}
\end{small}
\end{table}

\begin{table}[!htbp]
\centering
\caption{Per-class C-Score for DenseNet201. E25 shows 0.000 for all Normal-class
techniques (gold-list population effect, not method failure): no Normal images pass
$\tau=0.5$ at this checkpoint during boundary reorganisation.}
\label{tab:perclass_densenet}
\begin{small}
\setlength{\tabcolsep}{5pt}
\begin{tabular}{lccccccc}
\toprule
\textbf{Technique} & \textbf{E1} & \textbf{E5} & \textbf{E10} & \textbf{E15}
  & \makecell{\textbf{E20}\\\small{(TL)}}
  & \makecell{\textbf{E25}\\\small{(FT)}}
  & \makecell{\textbf{E30}\\\small{(FT)}} \\
\midrule
\multicolumn{8}{l}{\textit{Normal (Class 0)}} \\
\midrule
GradCAM      & 0.159 & 0.593 & 0.606 & 0.663 & 0.664 & \cd{0.000} & \cg{0.924} \\
GradCAM++    & 0.424 & 0.669 & 0.694 & 0.728 & 0.718 & \cd{0.000} & \cg{0.918} \\
LayerCAM     & 0.471 & 0.672 & 0.696 & 0.729 & 0.716 & \cd{0.000} & \cg{0.919} \\
ScoreCAM     & 0.539 & 0.621 & 0.688 & 0.738 & 0.714 & \cd{0.000} & \cg{0.907} \\
EigenCAM     & 0.680 & 0.682 & 0.688 & 0.689 & 0.688 & \cd{0.000} & \cg{0.895} \\
MS-GradCAM++ & 0.370 & 0.504 & 0.535 & 0.586 & 0.573 & \cd{0.000} & 0.674 \\
\midrule
\multicolumn{8}{l}{\textit{Pneumonia (Class 1)}} \\
\midrule
GradCAM      & 0.078 & 0.007 & 0.002 & 0.004 & 0.014 & 0.744 & 0.761 \\
GradCAM++    & 0.310 & 0.382 & 0.490 & 0.498 & 0.483 & \cg{0.916} & \cg{0.851} \\
LayerCAM     & 0.352 & 0.406 & 0.520 & 0.522 & 0.498 & \cg{0.915} & \cg{0.852} \\
ScoreCAM     & 0.401 & 0.541 & 0.597 & 0.587 & 0.597 & \cg{0.933} & \cg{0.869} \\
EigenCAM     & 0.603 & 0.615 & 0.615 & 0.614 & 0.614 & \cg{0.908} & \cg{0.826} \\
MS-GradCAM++ & 0.286 & 0.333 & 0.402 & 0.408 & 0.395 & 0.644 & 0.596 \\
\bottomrule
\end{tabular}
\end{small}
\end{table}

\begin{table}[!htbp]
\centering
\caption{Per-class C-Score for InceptionV3. GradCAM Pneumonia achieves 0.887 at E25
but collapses to 0.008 at E30, reflecting non-monotonic attribution dynamics post fine-tuning.}
\label{tab:perclass_inceptionv3}
\begin{small}
\setlength{\tabcolsep}{5pt}
\begin{tabular}{lccccccc}
\toprule
\textbf{Technique} & \textbf{E1} & \textbf{E5} & \textbf{E10} & \textbf{E15}
  & \makecell{\textbf{E20}\\\small{(TL)}}
  & \makecell{\textbf{E25}\\\small{(FT)}}
  & \makecell{\textbf{E30}\\\small{(FT)}} \\
\midrule
\multicolumn{8}{l}{\textit{Normal (Class 0)}} \\
\midrule
GradCAM      & 0.047 & 0.288 & 0.380 & 0.317 & 0.335 & \cg{0.840} & \cg{0.847} \\
GradCAM++    & 0.469 & 0.542 & 0.536 & 0.516 & 0.509 & \cg{0.872} & \cg{0.851} \\
LayerCAM     & 0.613 & 0.642 & 0.631 & 0.620 & 0.623 & \cg{0.866} & \cg{0.852} \\
ScoreCAM     & 0.292 & 0.332 & 0.303 & 0.286 & 0.286 & \cg{0.845} & \cg{0.852} \\
EigenCAM     & 0.770 & 0.777 & 0.775 & 0.774 & 0.774 & \cg{0.938} & \cg{0.922} \\
MS-GradCAM++ & 0.393 & 0.407 & 0.407 & 0.400 & 0.396 & 0.753 & 0.759 \\
\midrule
\multicolumn{8}{l}{\textit{Pneumonia (Class 1)}} \\
\midrule
GradCAM      & 0.244 & 0.218 & 0.138 & 0.146 & 0.140 & \cg{0.887} & 0.008 \\
GradCAM++    & 0.479 & 0.491 & 0.474 & 0.471 & 0.475 & 0.785 & 0.728 \\
LayerCAM     & 0.539 & 0.553 & 0.548 & 0.546 & 0.546 & 0.779 & 0.728 \\
ScoreCAM     & 0.453 & 0.414 & 0.416 & 0.419 & 0.417 & 0.770 & 0.723 \\
EigenCAM     & 0.750 & 0.750 & 0.750 & 0.750 & 0.749 & \cg{0.881} & \cg{0.825} \\
MS-GradCAM++ & 0.435 & 0.422 & 0.421 & 0.421 & 0.429 & 0.626 & 0.613 \\
\bottomrule
\end{tabular}
\end{small}
\end{table}

\begin{table}[!htbp]
\centering
\caption{Per-class C-Score for ResNet50V2. Normal class registers 0.000 across all
techniques at E30 (mode collapse). Pneumonia GradCAM at E25 drops to 0.000 while
GradCAM++ and LayerCAM remain above 0.600.}
\label{tab:perclass_resnet}
\begin{small}
\setlength{\tabcolsep}{5pt}
\begin{tabular}{lccccccc}
\toprule
\textbf{Technique} & \textbf{E1} & \textbf{E5} & \textbf{E10} & \textbf{E15}
  & \makecell{\textbf{E20}\\\small{(TL)}}
  & \makecell{\textbf{E25}\\\small{(FT)}}
  & \makecell{\textbf{E30}\\\small{(FT)}} \\
\midrule
\multicolumn{8}{l}{\textit{Normal (Class 0)}} \\
\midrule
GradCAM      & 0.333 & 0.095 & 0.573 & 0.455 & 0.481 & 0.798 & \cd{0.000} \\
GradCAM++    & 0.364 & 0.656 & 0.662 & 0.643 & 0.634 & 0.798 & \cd{0.000} \\
LayerCAM     & 0.517 & 0.715 & 0.711 & 0.696 & 0.688 & 0.798 & \cd{0.000} \\
ScoreCAM     & 0.508 & 0.882 & 0.652 & 0.645 & 0.649 & 0.041 & \cd{0.000} \\
EigenCAM     & 0.593 & 0.745 & 0.727 & 0.728 & 0.723 & 0.802 & \cd{0.000} \\
MS-GradCAM++ & 0.333 & 0.495 & 0.518 & 0.508 & 0.508 & 0.595 & \cd{0.000} \\
\midrule
\multicolumn{8}{l}{\textit{Pneumonia (Class 1)}} \\
\midrule
GradCAM      & 0.463 & 0.493 & 0.572 & 0.379 & 0.348 & \cd{0.000} & 0.370 \\
GradCAM++    & 0.300 & 0.598 & 0.635 & 0.593 & 0.578 & 0.613 & 0.478 \\
LayerCAM     & 0.511 & 0.650 & 0.669 & 0.648 & 0.640 & 0.611 & 0.489 \\
ScoreCAM     & 0.521 & 0.632 & 0.621 & 0.611 & 0.597 & \cd{0.000} & \cd{0.000} \\
EigenCAM     & 0.587 & 0.664 & 0.680 & 0.678 & 0.675 & 0.614 & 0.495 \\
MS-GradCAM++ & 0.303 & 0.512 & 0.530 & 0.507 & 0.507 & 0.500 & 0.409 \\
\bottomrule
\end{tabular}
\end{small}
\end{table}

\end{document}